\documentclass[11pt]{article}

\usepackage[preprint]{acl}

\usepackage{times}
\usepackage{latexsym}
\usepackage[T1]{fontenc}
\usepackage[utf8]{inputenc}
\usepackage{microtype}
\usepackage{inconsolata}
\usepackage{graphicx}
\usepackage{booktabs}
\usepackage{multirow}
\usepackage{amsmath}
\usepackage{xcolor}
\usepackage{pgfplots}
\pgfplotsset{compat=1.18}
\usetikzlibrary{arrows.meta}

\graphicspath{{figures/}}

\newcommand{\Nstar}{$N^\ast$}

\title{Coverage Governs Accuracy: When to Expand the Lexicon\\and When to Fall Back in Low-Resource Manchu OCR}

\author{
  Ruixin Yu\textsuperscript{1},
  Zifan Wang\textsuperscript{1},
  Mingxin Ding\textsuperscript{1},
  Xiaoxin Sun\textsuperscript{1}\thanks{Corresponding author.}, \\
  \textbf{Dongjie Wang\textsuperscript{1},
  Suhua Wang\textsuperscript{2},
  Minghao Yin\textsuperscript{1}} \\[4pt]
  \textsuperscript{1}School of Information Science and Technology, Northeast Normal University \\
  \textsuperscript{2}Computer Department, Changchun Humanities and Sciences College \\
  \texttt{yuruixin114@nenu.edu.cn}, \texttt{sunxx772@nenu.edu.cn}
}

\begin{document}
\maketitle

\begin{abstract}
Character-level OCR for low-resource Manchu often produces near-miss spelling errors. Lexicon-constrained decoding can correct such errors with lexical knowledge, but a fixed dictionary cannot cover the many unseen word forms produced by agglutinative word formation. We study whether morphological expansion alleviates this limitation, and how the expansion size and confidence fallback affect recognition. We segment the training transcriptions with a dictionary-based stem list, extract roots and suffix chains from them, generate about 510k candidates, select the expansion size through frequency ranking on the validation set, and decompose accuracy under hard lexicon constraints into the product of coverage and in-lexicon precision. Across three CRNN training seeds, adding 250k candidates to a base dictionary of about 29k words raises the type accuracy on unseen combinations of seen roots from 63.6\% to 86.3\%, and overall token accuracy from 82.4\% to 84.2\%. Adding all candidates lowers overall token accuracy to 81.5\%, showing that the coverage gain comes with a loss of in-lexicon precision. A further comparison shows that the base dictionary with confidence fallback performs best overall among the CRNN systems tested, reaching 76.1\% type and 85.5\% token accuracy and outperforming the expanded lexicon both with and without fallback. A Transformer recognizer shows similar trends. These results indicate that morphological expansion relieves the constraint of a fixed dictionary, but a larger lexicon does not guarantee better recognition; in the present low-resource Manchu OCR setting, a selective combination of reliable lexical constraints and open recognition is more effective.
\end{abstract}

\section{Introduction}

Large collections of Manchu archives and books record the history, administration and social life of the Qing dynasty, but the text in their scanned images must be converted into searchable and collatable text before it can support further organization and research. Optical character recognition (OCR) is the basis of this process. The vertical, cursive script of Manchu and the limited amount of annotated data make reliable recognition difficult. Under low-resource conditions, the outputs of character-level recognizers are often close to the correct answer but contain a few spelling errors: on our test set, 73--78\% of the greedy errors on the two main generalization tiers are a single edit away from the correct word. How to bring in usable linguistic knowledge beyond the limited training data is therefore an important question for improving Manchu OCR.

Lexicon-constrained decoding is a direct way to use such knowledge. Restricting decoding to a lexicon rules out some implausible strings and lets the system reselect among candidate words on the basis of image evidence. The lexicon thus provides an explicit lexical prior, so that recognition no longer relies entirely on character-by-character prediction. The constraint, however, has a clear boundary: the system can output the correct word only if the lexicon already contains it.

This boundary is particularly pronounced in agglutinative languages such as Manchu. Manchu builds different word forms by attaching one or more suffixes to a stem, and some suffixes vary with the vowel-harmony class of the stem. For example, the perfective participle suffix has the forms \textit{-ha}, \textit{-he} and \textit{-ho}: \textit{ara-} `write' with \textit{-ha} gives \textit{araha} `wrote', and \textit{gene-} `go' with \textit{-he} gives \textit{genehe} `went'. \textit{araha} can be further followed by \textit{-ngge}, giving \textit{arahangge} `that which was written'. Stems, suffix variants and their multi-layer combinations together widen the range of word forms found in real documents \citep{gorelova2002manchu,mollendorff1892manchu}. Moreover, some affixes in Qing-dynasty Manchu documents can be written either joined to the stem or separately, so morphological membership does not always coincide with written word boundaries. A fixed dictionary can therefore hardly exhaust the word-form combinations in real documents, and its coverage in word-level OCR also depends on the transliteration and segmentation conventions of its entries and of the image labels. In our test set, only 0.8\% of word images correspond to a form seen in training. Even when the root has been seen, a new root--suffix combination may be missing from the base dictionary. Based on this observation, we segment the training transcriptions using the base dictionary, extract roots and suffix chains from them, and generate and filter candidate word forms, to examine how far morphological expansion can make up for the limited coverage of a fixed dictionary.

Does expanding the lexicon, however, necessarily keep improving recognition accuracy? New candidates may add correct answers, but they also add word forms that compete with the correct ones. For inputs the base dictionary already covers, more candidates may bring no further benefit and may instead lead the recognizer to a wrong result. The coverage gain of expansion and the loss of in-lexicon precision therefore need to be measured together. Furthermore, when the correct word is probably not in the lexicon, should the system still enforce the lexicon constraint, or should it retain the ability of the character-level recognizer to output unlisted forms? Together, these two questions determine at what scale, and in what way, lexical knowledge should take part in low-resource OCR.

\begin{figure}[t]
\centering
\begin{tikzpicture}
\begin{axis}[
  width=\columnwidth, height=0.78\columnwidth,
  xmin=89.5, xmax=102.4, ymin=78.5, ymax=92,
  xtick={90,92,94,96,98,100},
  xlabel={Lexicon coverage of test images (\%)},
  ylabel={In-lexicon precision (\%)},
  xlabel style={font=\small}, ylabel style={font=\small},
  tick label style={font=\footnotesize},
  grid=major, grid style={gray!15},
  axis lines=left,
  clip=true,
]
\foreach \a in {80,82.5,85,87.5} {
  \addplot[gray!55, thin, dashed, domain=89.5:101, samples=40] {100*\a/x};
}
\node[font=\scriptsize, gray, anchor=west, fill=white, inner sep=0.5pt] at (axis cs:100.95,78.9) {80};
\node[font=\scriptsize, gray, anchor=west, fill=white, inner sep=0.5pt] at (axis cs:100.95,81.4) {82.5};
\node[font=\scriptsize, gray, anchor=west, fill=white, inner sep=0.5pt] at (axis cs:100.95,83.9) {85};
\node[font=\scriptsize, gray, anchor=west, fill=white, inner sep=0.5pt] at (axis cs:100.95,86.3) {87.5};
\draw[-{Latex[length=2mm]}, thick, black!70] (axis cs:90.85,90.70) -- (axis cs:94.45,89.10);
\draw[-{Latex[length=2mm]}, thick, black!70, densely dotted] (axis cs:94.62,88.85) -- (axis cs:94.80,86.25);
\draw[-{Latex[length=2mm]}, very thick, black] (axis cs:91.1,90.5) -- (axis cs:99.7,85.65);
\addplot[only marks, mark=*, mark size=2.4pt, black] coordinates {(90.85,90.70) (94.58,89.00) (94.80,86.01)};
\addplot[only marks, mark=square*, mark size=2.4pt, black] coordinates {(100,85.50) (100,82.59)};
\addplot[only marks, mark=square, mark size=2.4pt, black] coordinates {(100,79.81) (100,80.32)};
\node[font=\scriptsize, anchor=south] at (axis cs:90.85,90.95) {Dictionary};
\node[font=\scriptsize, anchor=south west] at (axis cs:94.62,89.05) {+ top-$N^\ast$};
\node[font=\scriptsize, anchor=north east] at (axis cs:94.75,85.95) {Full union};
\node[font=\scriptsize, anchor=east] at (axis cs:99.6,85.5) {\textbf{Dict.\ + fallback}};
\node[font=\scriptsize, anchor=east] at (axis cs:99.6,82.59) {$N^\ast$ + fallback};
\node[font=\scriptsize, anchor=east] at (axis cs:99.6,80.6) {Open beam + LM};
\node[font=\scriptsize, anchor=east] at (axis cs:99.6,79.5) {Greedy};
\node[font=\scriptsize, anchor=south, rotate=-20] at (axis cs:96.2,87.75) {fallback};
\node[font=\scriptsize, anchor=north, rotate=-24] at (axis cs:92.5,89.75) {expansion};
\end{axis}
\end{tikzpicture}
\caption{Systems on the coverage--precision plane (token level, all 5{,}389 test images, mean of three CRNN seeds). Dashed curves mark equal accuracy (coverage $\times$ precision). Adding 250k generated words (\Nstar) or all 510k candidates (full union) to the dictionary increases coverage but lowers precision. Confidence fallback reaches full coverage and lies on a higher accuracy contour. Open-decoding systems have full coverage by construction.}
\label{fig:overview}
\end{figure}
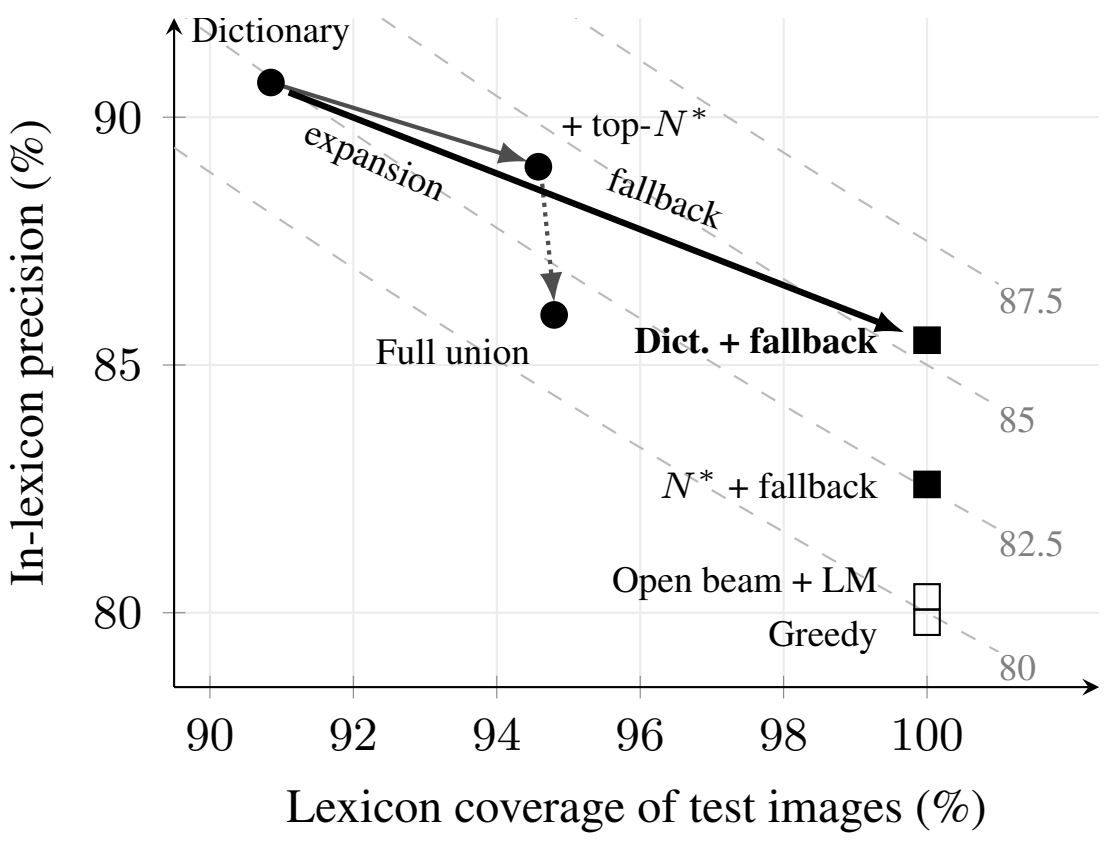

In this work, we address these questions by comparing morphological expansion with confidence fallback. We first decompose accuracy under hard lexicon constraints exactly into the product of lexicon coverage and in-lexicon precision, which separates whether the correct answer is in the candidate space from whether the recognizer can select it from the candidates (Figure~\ref{fig:overview}). Starting from a base lexicon of 29{,}024 words, we then generate about 510k candidates from the roots and suffix chains in the training data, rank them by frequency and control how many of them enter the lexicon. In parallel, we introduce confidence fallback, which compares the CTC \citep{graves2006ctc} scores of the constrained result and the greedy result and adopts the open output when the constrained result lacks sufficient support. Both the lexicon size and the fallback threshold are selected on the validation set. The evaluation is stratified into seen words, unseen combinations of seen roots, unseen roots and unanalysable forms, to examine gains and losses under different lexical conditions.

Morphological expansion does relieve the constraint of a fixed dictionary, but closer experiments reveal the limits of its benefit. Across three CRNN training seeds, adding 250k ranked candidates raises the type accuracy on unseen combinations of seen roots from 63.6\% to 86.3\%, and token accuracy on the full test set from 82.4\% to 84.2\%. Adding all candidates, however, lowers overall token accuracy to 81.5\%, and the drop is especially pronounced on the unseen-root tier, where the base dictionary already covers the vast majority of images. Confidence fallback can alleviate insufficient coverage, but combining it with lexicon expansion does not necessarily help: among the CRNN systems, the base dictionary with fallback achieves the highest overall type accuracy of 76.1\% and token accuracy of 85.5\%, outperforming the 250k-candidate lexicon both with and without fallback. Even under oracle fallback, which routes each image according to whether its correct word is in the lexicon, the base dictionary still reaches a higher token accuracy than the expanded lexicon. These results suggest that, in the present low-resource Manchu OCR setting, raising lexical coverage also requires controlling the precision cost of the constraint, and that a selective combination of a reliable lexicon and open recognition is more effective than simply enlarging the candidate space.

Our main contributions are as follows:
\begin{itemize}
\item We build and evaluate a morphology-driven lexicon expansion scheme. Using the dictionary as the basis for segmentation, we extract roots and suffix chains from the training transcriptions, generate about 510k candidates, and control the expansion size through frequency ranking and validation-set selection, to alleviate the limited coverage of a fixed dictionary for unseen word forms.
\item We reveal the benefits and limits of lexicon expansion. Through the coverage--in-lexicon precision decomposition and controlled comparisons of candidate size, ranking method and recognizer, we show that higher coverage comes with a loss of in-lexicon precision, and that a larger expansion is not always better.
\item We demonstrate the complementary value of lexical constraints and open recognition. Comparing combinations of lexicons and confidence fallback, we find that the base dictionary with fallback performs best overall in our setting, which provides empirical guidance on the use of lexical constraints in low-resource Manchu OCR.
\end{itemize}

\section{Background and Related Work}

\paragraph{Lexicon constraints in text recognition.} Word lexicons have long been used to constrain the output of handwriting and scene-text recognizers. One line of work restricts CTC decoding to a closed lexicon \citep{graves2009handwriting}, while another keeps decoding open and adds a character or word language model through prefix beam search \citep{hannun2014firstpass}. Attention-based recognizers such as TrOCR \citep{li2023trocr} and PARSeq \citep{bautista2022parseq} learn an implicit language model from their training text, which is of limited help when most test words are unseen. Our work compares closed and open decoding in such a setting and quantifies where a closed lexicon loses accuracy.

\paragraph{Manchu and related scripts.} Earlier work on Manchu OCR has addressed character segmentation and recognition \citep{xu2016manchu}, segmentation-free recognition of printed words with synthetic data and convolutional features \citep{huang2017manchu,zheng2018manchu}, handwritten word recognition with pretrained and Transformer models \citep{li2023manchu,guo2025manchuswin}, and the restoration of damaged word images \citep{bi2026rgca}. \citet{zhang2005manchu} used language-model post-processing for handwritten Manchu characters. For traditional Mongolian, a related vertical script, sub-word units \citep{fan2019mongolian} and multi-task learning \citep{wei2021mongolian} have been applied to word recognition. More broadly, \citet{rijhwani2020ocrpost} showed that post-correction can substantially reduce OCR errors on endangered-language texts. None of these studies separates lexicon coverage from in-lexicon precision or compares lexicon expansion with fallback to open decoding.

\paragraph{Open vocabularies and morphology.} Text models commonly handle open vocabularies through sub-word units \citep{sennrich2016bpe} or unsupervised morphological segmentation \citep{creutz2007morfessor}. We instead generate whole-word candidates from roots and suffix chains. This choice keeps decoding at the word level, where near misses can be corrected, and makes lexicon coverage directly measurable.

\section{Method}
\label{sec:method}

\subsection{Problem Formulation}

Given the image of a single word, a recognizer trained with CTC produces a distribution over character sequences. Greedy decoding takes the most likely label in each frame, and open beam search explores several character sequences, optionally scored by a character language model. Lexicon-constrained beam search instead keeps only hypotheses that are prefixes of words in a lexicon $V$, so that the output is always an element of $V$. We are interested in how the choice of $V$, and the decision whether to constrain at all, affects recognition accuracy.

\subsection{Coverage--Precision Decomposition}
\label{sec:decomp}

Let $T$ be the set of word types in the test set and $\mathrm{acc}(t)$ the fraction of images of type $t$ that are recognized correctly. Since a constrained decoder can only output words in $V$, every type outside $V$ has $\mathrm{acc}(t)=0$, and type accuracy factorizes exactly:
\begin{equation}
\mathrm{Acc} = \underbrace{\frac{|T\cap V|}{|T|}}_{\text{coverage}} \times \underbrace{\frac{1}{|T\cap V|}\sum_{t\in T\cap V}\mathrm{acc}(t)}_{\text{in-lexicon precision}}.
\label{eq:decomp}
\end{equation}
The same identity holds for token accuracy when images take the place of types. The two factors have different sources. Coverage is a property of the lexicon and the test labels alone, whereas precision depends on the recognizer and on how many plausible competitors the lexicon offers for each correct word. The decomposition therefore predicts that changing the recognizer leaves coverage unchanged, and that enlarging the lexicon can raise coverage while lowering precision. We test both predictions in \S\ref{sec:analysis}.

\subsection{Lexicon Expansion}

Our base lexicon is the word list of a dictionary-based morphological analyzer, which we refer to as the \emph{dictionary}. It contains 29{,}024 entries: 23{,}277 derived from external dictionary resources, including the headwords of the electronic New Manchu--Chinese Dictionary \citep{hu2020dict}, and 5{,}747 word forms added from an external corpus of Manchu archival texts. To cover word forms the dictionary lacks, we segment the training words into roots and suffix chains with the analyzer, collect 2{,}879 roots and 181 suffix chains, and pair every root with every chain. The 521{,}099 pairs yield 508{,}588 distinct surface forms; together with 1{,}574 training word forms that concatenation does not produce, this gives 510{,}162 candidates. We score each candidate as $P(\text{root})\,P(\text{chain})$, with both probabilities estimated from frequencies in the training images, and add the $N$ highest-scoring candidates to the dictionary. We select $N$ on the validation set (\S\ref{sec:setup}), which gives \Nstar{} $=250{,}000$; adding all candidates gives the \emph{full union} of 530{,}025 words.

\subsection{Confidence Fallback}

Fallback obtains coverage without changing the lexicon. Each image is decoded twice, once with lexicon-constrained beam search and once greedily. If the score of the best constrained hypothesis falls more than a threshold $\tau$ below the score of the greedy string, the system outputs the greedy string, and otherwise the constrained one. The output can only change when the greedy string lies outside the lexicon, so the rule effectively acts as a detector for inputs the lexicon does not cover. We select $\tau$ on the validation set.

\section{Experimental Setup}
\label{sec:setup}

\paragraph{Data.} Each example is a cropped image of one vertically written Manchu word, transcribed with the Latin transliteration character set of Manchu. The images come from four previously published sources (Appendix~\ref{app:data}): the printed Manchu50000 collection and three handwritten collections, MengGuTangDang, SuiJunJiXing and the Grand Council (Junjichu) Manchu archives. Manchu experts segmented the word images, supplied the transliterations and reviewed their quality; ambiguous samples were resolved through review, and samples that could not be identified reliably were excluded. The training set contains 55{,}240 images of 6{,}485 word types, the validation set 4{,}767 images, and the test set 5{,}389 images of 792 word types. All four sources occur in every split, since the split holds out word forms rather than source documents.

\paragraph{Test tiers.} To measure generalization beyond the training vocabulary, we assign each test word to one of four tiers according to its morphological analysis (Table~\ref{tab:tiers}). Tier~A contains words that occur in training. Tier~B contains words whose root occurs in training but whose root--suffix combination does not; within B, we further distinguish forms whose suffix chain was seen in training (B1) from forms with an unseen chain (B2). Tier~C contains words whose root never occurs in training, and words that the segmenter cannot analyse form a separate unresolved tier. Tiers B and C account for 95\% of the test images, and we report results for all tiers as well as for the full test set.

\begin{table}[t]
\centering\small
\begin{tabular}{llrr}
\toprule
Tier & Definition & Types & Images \\
\midrule
A & word seen in training & 34 & 41 \\
B & root seen, combination unseen & 336 & 2{,}542 \\
C & root unseen & 264 & 2{,}572 \\
-- & unresolved by segmenter & 158 & 234 \\
\midrule
All & & 792 & 5{,}389 \\
\bottomrule
\end{tabular}
\caption{Test tiers. Only tier A (0.8\% of images) contains words seen in training.}
\label{tab:tiers}
\end{table}

\paragraph{Recognizers.} Our main recognizer is a CRNN trained with CTC \citep{shi2017crnn} with 14.3M parameters. To test whether our findings depend on the architecture, we also train a Transformer recognizer with 19.6M parameters. Both models are trained for 30 epochs with a batch size of 384 using AdamW \citep{loshchilov2019adamw} with a learning rate of $10^{-3}$, weight decay of $10^{-4}$, one warm-up epoch, cosine decay to 2\% of the peak rate, gradient clipping at 5.0 and bf16 precision, without data augmentation. We keep the checkpoint with the best validation type accuracy. We train the CRNN with three seeds (42, 43 and 44) and report the mean and sample standard deviation; the Transformer is trained with one seed. Several analyses in \S\ref{sec:analysis} use the \emph{original model}, an earlier exploratory CRNN run with the same architecture and hyperparameters that was trained with data parallelism on three GPUs (128 images per GPU, the same effective batch size); all decoding experiments were first run on this model, so its numbers differ slightly from the three-seed means in Table~\ref{tab:main}.

\paragraph{Decoding.} All beam searches use a beam width of 50; on tiers B and C, widths of 10, 50 and 100 change accuracy by at most 0.1 points. For open beam search, we optionally add a character trigram language model trained on the training transcriptions, with weight $\alpha=0.25$ and length bonus $\beta=0.25$ \citep{hannun2014firstpass}. We also test two variants of the full union: a conservative vowel-harmony filter that removes 40{,}439 candidates, and a soft frequency prior added to the beam score with weight $\lambda=0.01$. All decoding hyperparameters ($N$, $\tau$, $\alpha$, $\beta$ and $\lambda$) are selected on validation scores only, as are the recognizer checkpoints. The test set is not entirely untouched, however: it was decoded in earlier exploratory experiments, and accuracy on tier A was logged during training.

\paragraph{Evaluation.} We report two exact-match metrics. Token accuracy is the fraction of test images whose word is recognized correctly. Type accuracy averages the per-type image accuracy over word types with equal weight, so that frequent words do not dominate the score. We compute 95\% confidence intervals with a cluster bootstrap over word types (10{,}000 resamples, keeping all images of a type together; \citealp{efron1993bootstrap}), and compare systems with exact McNemar tests \citep{mcnemar1947} and cluster sign-flip tests with Holm correction \citep{holm1979}.

\section{Main Results}
\label{sec:results}

Table~\ref{tab:main} summarizes the results of all systems. We organize the discussion around the three groups of systems in the table.

\begin{table*}[t]
\centering\small
\resizebox{\textwidth}{!}{%
\begin{tabular}{lrrrrrrrr}
\toprule
 & \multicolumn{2}{c}{Tier B (seen root)} & \multicolumn{2}{c}{Tier C (unseen root)} & \multicolumn{2}{c}{All, CRNN (3 seeds)} & \multicolumn{2}{c}{All, Transformer} \\
System & Type $\uparrow$ & Token $\uparrow$ & Type $\uparrow$ & Token $\uparrow$ & Type $\uparrow$ & Token $\uparrow$ & Type $\uparrow$ & Token $\uparrow$ \\
\midrule
\multicolumn{9}{l}{\textit{Open decoding (coverage 100\% by construction)}} \\
Greedy CTC & 83.5$\pm$0.5 & 85.4$\pm$0.4 & 78.1$\pm$2.9 & 77.3$\pm$0.9 & 74.0$\pm$1.1 & 79.8$\pm$0.3 & 71.9 & 74.6 \\
Open beam & 83.7$\pm$0.4 & 85.4$\pm$0.4 & 78.2$\pm$2.8 & 77.7$\pm$0.8 & 74.2$\pm$1.3 & 80.0$\pm$0.4 & 71.9 & 74.7 \\
Open beam + char.\ LM & 84.4$\pm$0.6 & 87.2$\pm$0.2 & 77.7$\pm$3.1 & 76.6$\pm$1.4 & 74.2$\pm$1.5 & 80.3$\pm$0.6 & 72.8 & 75.4 \\
\midrule
\multicolumn{9}{l}{\textit{Lexicon-constrained decoding}} \\
Dictionary (29k) & 63.6$\pm$0.2 & 84.0$\pm$0.8 & 75.0$\pm$1.0 & \textbf{88.0}$\pm$0.4 & 56.2$\pm$0.4 & 82.4$\pm$0.5 & 55.9 & 79.9 \\
Dictionary + top-\Nstar{} (250k) & \textbf{86.3}$\pm$0.5 & \textbf{89.0}$\pm$0.6 & 76.2$\pm$1.1 & 86.8$\pm$0.4 & 66.5$\pm$0.7 & 84.2$\pm$0.3 & 65.5 & \textbf{81.0} \\
Full union (530k) & 85.9$\pm$0.6 & 88.7$\pm$0.6 & 76.5$\pm$1.4 & 81.5$\pm$1.1 & 66.5$\pm$0.8 & 81.5$\pm$0.3 & 65.6 & 77.6 \\
\quad + vowel-harmony filter & 86.0$\pm$0.6 & 88.7$\pm$0.6 & 76.6$\pm$1.3 & 81.6$\pm$1.1 & 66.6$\pm$0.8 & 81.6$\pm$0.3 & 65.6 & 77.6 \\
\quad + frequency prior & 85.9$\pm$0.7 & 88.7$\pm$0.6 & 76.5$\pm$1.3 & 81.4$\pm$1.2 & 66.5$\pm$0.7 & 81.5$\pm$0.3 & 65.6 & 77.6 \\
\midrule
\multicolumn{9}{l}{\textit{Lexicon-constrained decoding with confidence fallback}} \\
\Nstar{} + fallback & 85.4$\pm$0.3 & 87.5$\pm$0.8 & 80.5$\pm$2.7 & 81.2$\pm$0.5 & 75.6$\pm$1.2 & 82.6$\pm$0.2 & 73.9 & 77.5 \\
Dictionary + fallback & 85.3$\pm$0.6 & \textbf{89.0}$\pm$0.8 & \textbf{83.8}$\pm$1.7 & 86.1$\pm$0.5 & \textbf{76.1}$\pm$0.9 & \textbf{85.5}$\pm$0.5 & \textbf{75.4} & 80.9 \\
\bottomrule
\end{tabular}}
\caption{Test accuracy (\%). CRNN columns show the mean $\pm$ sample standard deviation over three training seeds; the Transformer was trained with one seed. ``All'' includes tier A and unresolved words. Decoding hyperparameters are tuned on the validation set. The best value in each column is in bold.}
\label{tab:main}
\end{table*}

\paragraph{Open versus constrained decoding.} Greedy CTC decoding reaches 74.0\% type and 79.8\% token accuracy over all images. Its errors are typically near misses: among erroneous greedy outputs on tiers B and C, 73--78\% are one edit away from the gold word and 64--81\% are not valid words (Appendix Figure~\ref{fig:errors}). A character language model corrects few of them, raising token accuracy by only 0.3 points. Constraining the output to the dictionary has a very different effect on the two tiers. On tier C, token accuracy increases from 77.3\% to 88.0\%, because the dictionary covers most of these unseen-root words and corrects their near misses. On tier B, by contrast, type accuracy drops to 63.6\%, since the dictionary contains only 68.5\% of the tier-B word types. Once decoding is constrained, the lexicon rather than the recognizer limits accuracy.

\paragraph{Lexicon expansion.} Adding the \Nstar{} highest-ranked combinations to the dictionary resolves the coverage problem on tier B, where type accuracy rises from 63.6\% to 86.3\%. The same expansion brings little benefit on tier C, where generated candidates cannot reproduce unseen roots: token accuracy falls slightly from 88.0\% to 86.8\%, and drops to 81.5\% when all 510k candidates are added. Neither the vowel-harmony filter nor the soft frequency prior changes this outcome. Over all images, \Nstar{} improves token accuracy by 1.8 points over the dictionary, but its type accuracy (66.5\%) remains well below that of greedy decoding, because the unresolved words and part of tier C stay outside the lexicon.

\paragraph{Confidence fallback.} Adding fallback to the dictionary gives the best CRNN system on both metrics, with 76.1\% type and 85.5\% token accuracy over all images. The gain is concentrated on tier C, where type accuracy reaches 83.8\%, compared with 78.1\% for greedy decoding and 75.0\% for the dictionary alone, while tier-B token accuracy matches that of \Nstar. On the original model, fallback improves the dictionary by 19.0 type points (95\% CI [16.2, 21.9]) and 3.0 token points ([1.3, 5.4]). In contrast, combining fallback with \Nstar{} is less effective and lowers tier-C token accuracy by 5.6 points relative to \Nstar{} alone; we analyse the reason in \S\ref{sec:routing}.

\paragraph{Transformer recognizer.} The Transformer is weaker than the CRNN overall, with 71.9\% type and 74.6\% token accuracy under greedy decoding, but the relative ordering of the decoding strategies is largely the same. The dictionary with fallback again achieves the highest type accuracy (75.4\%), and its token accuracy is within 0.1 points of \Nstar.

\section{Analysis}
\label{sec:analysis}

The main results show that fallback is a better route to coverage than expansion. In this section, we use the decomposition in Eq.~\ref{eq:decomp} to examine why.

\subsection{Coverage and Precision Are Separable}

The decomposition predicts that coverage depends only on the lexicon, whereas precision also depends on the recognizer. To test this, we train CRNNs on nested subsets containing 25\%, 50\% and 100\% of the training images and decode them with the same lexicons. As expected, coverage is identical across the three models for every lexicon and tier. In-lexicon precision, in contrast, increases with the amount of training data for every lexicon (Table~\ref{tab:cross}). Larger lexicons have lower precision at every training size, but the gap between the dictionary and \Nstar{} narrows from 6.5 to 3.9 points as the training data grows. A stronger recognizer is therefore better at rejecting the additional competitors, which makes lexicon expansion more attractive as recognition quality improves.

\begin{table}[t]
\centering\small
\begin{tabular}{llrrr}
\toprule
 & Lexicon & 25\% & 50\% & 100\% \\
\midrule
\multirow{2}{*}{B} & Dictionary & 91.0 & 92.0 & 92.6 \\
 & \Nstar & 82.7 & 88.2 & 88.6 \\
\midrule
\multirow{2}{*}{C} & Dictionary & 85.5 & 88.8 & 91.0 \\
 & \Nstar & 82.1 & 86.9 & 89.4 \\
\midrule
\multirow{3}{*}{All} & Dictionary & 88.6 & 90.8 & 92.4 \\
 & \Nstar & 82.1 & 86.8 & 88.5 \\
 & Gap $\downarrow$ & 6.5 & 4.0 & 3.9 \\
\bottomrule
\end{tabular}
\caption{In-lexicon type precision (\%) of CRNNs trained on nested fractions of the training images (seed 42). Coverage does not change with the training fraction (dictionary: B 68.5, C 82.2; \Nstar: B 97.0, C 84.8).}
\label{tab:cross}
\end{table}

Across the 14 tier-B and tier-C lexicon configurations of the original model, type accuracy lies close to a line through the origin with a slope of 0.895 ($R^2=0.989$; Appendix Figure~\ref{fig:fit}). Since these points share the same model and test images, we treat the fit as descriptive. It nevertheless illustrates that, within this range of lexicons, differences in coverage account for most of the differences in accuracy.

\subsection{Sources of Coverage and the Choice of $N$}

\begin{figure*}[t]
\centering
\includegraphics[width=0.82\textwidth]{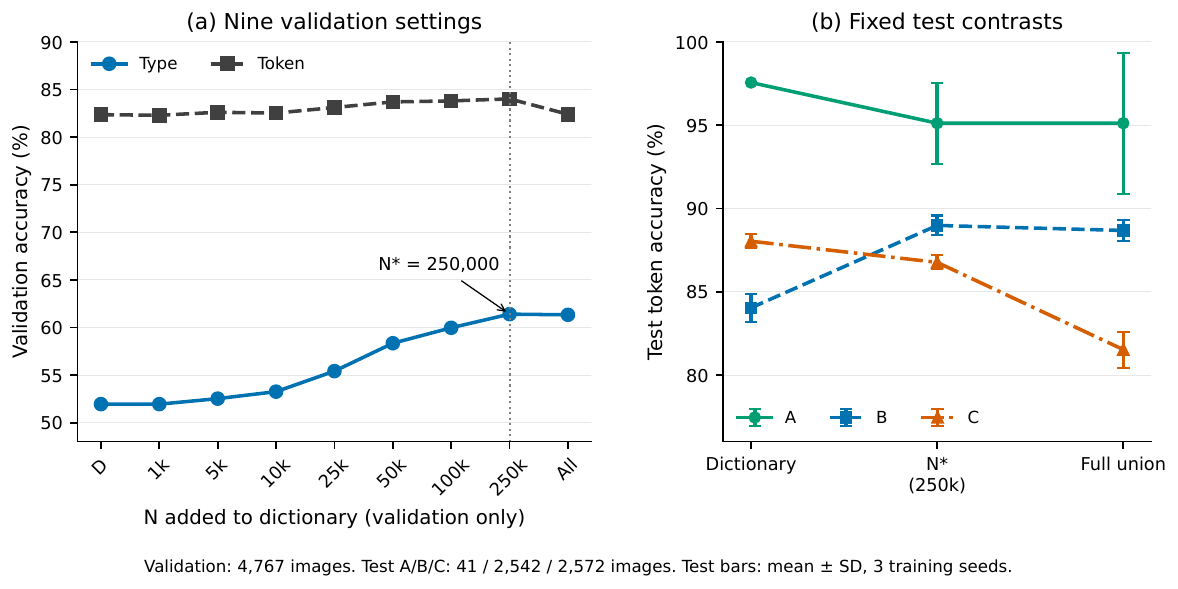}
\caption{Effect of the number $N$ of generated candidates. (a) Validation type and token accuracy as the top-$N$ candidates are added to the dictionary; \Nstar{} $=250$k maximizes type accuracy. (b) Test token accuracy per tier for the dictionary, \Nstar{} and the full union (three CRNN seeds, mean $\pm$ SD).}
\label{fig:scale}
\end{figure*}

Generated combinations and the dictionary cover largely different words. The combinations cover 96.7\% of tier-B types but only 19.7\% of tier-C types, whose roots are unseen by definition, while the dictionary covers 68.5\% and 82.2\%, respectively. Together they cover 97.6\% of tier B and 87.1\% of tier C. Figure~\ref{fig:scale} shows how accuracy changes as more ranked combinations are added. On the validation set, type accuracy increases steadily up to \Nstar{} $=250$k, whereas adding all 510k candidates lowers token accuracy from 84.0\% to 82.4\%. On the test set, the same expansion helps tier B and hurts tier C.

\subsection{Which Ranking Signal Matters?}

Frequency ranking is essential for expansion: at $N=250$k, it outperforms a random selection of the same size by 12.0 type points on tier B (95\% CI [9.5, 14.6]). To identify where this signal comes from, we compare alternative scoring rules at the same $N$ (Table~\ref{tab:rank}). Ranking candidates by suffix-chain frequency alone performs as well as the full root $\times$ chain score, whereas ranking by root frequency alone loses 8.3 type points on tier B. Restricting the lexicon to combinations attested in an external corpus performs no better than random selection. This result is consistent with the sparsity of the corpus, which contains only 1.08\% of the 510k candidates, and suggests that, in this low-resource setting, corpus attestation is too sparse to identify plausible word forms.

\begin{table}[t]
\centering\small
\setlength{\tabcolsep}{4pt}
\begin{tabular}{lrrrr}
\toprule
 & \multicolumn{2}{c}{Tier B} & \multicolumn{2}{c}{All} \\
Ranking ($N=250$k) & Type $\uparrow$ & Tok.\ $\uparrow$ & Type $\uparrow$ & Tok.\ $\uparrow$ \\
\midrule
Root $\times$ chain & 87.7 & 88.4 & \textbf{66.3} & \textbf{84.6} \\
Suffix chain only & \textbf{87.8} & \textbf{88.6} & 66.0 & \textbf{84.6} \\
Root only & 79.4 & 86.9 & 63.1 & 84.0 \\
Corpus-attested & 75.9 & 86.3 & 61.1 & 81.6 \\
Random (5 draws) & 75.7 & 85.6 & 61.3 & 81.9 \\
\bottomrule
\end{tabular}
\caption{Ranking rules for generated candidates (original model, accuracy \%).}
\label{tab:rank}
\end{table}

\subsection{The Precision Cost of Expansion}

The cost of expansion is most visible on tier C, where the dictionary already covers 97.6\% of the images. For these words, generated candidates add competitors rather than answers, and token accuracy falls from 88.0\% with the dictionary to 81.5\% with the full union. On the original model, in-lexicon type precision decreases by 3--4 points for every tenfold increase in lexicon size when candidates are ranked by frequency, but by only 1.5--2 points when they are added at random. One explanation is that frequency-ranked candidates more closely resemble real word forms and are therefore harder for the recognizer to reject. The direct evidence for this explanation is limited, however: the number of added edit-distance-1 neighbours of a word correlates only weakly with its accuracy loss (Spearman $\rho=0.13$ on tier C, exploratory $p=0.05$; not significant on tier B; Appendix Figure~\ref{fig:density}).

\subsection{Why Does Fallback Work?}
\label{sec:routing}

Since fallback can only change the output when the greedy string lies outside the lexicon, its effect depends on how well it identifies uncovered inputs. Table~\ref{tab:routing} analyses this routing decision on the original model. With the dictionary, fallback sends 76.1\% of the uncovered images to greedy decoding and misroutes only 3.4\% of the covered images. With \Nstar, the misrouting rate doubles to 7.7\%. In a 250k-word lexicon, an incorrect greedy string is more likely to be a valid entry by chance, which makes the score difference less informative and explains why fallback lowers the accuracy of \Nstar{} on tier C.

\begin{table}[t]
\centering\small
\begin{tabular}{lrr}
\toprule
 & Dictionary & \Nstar \\
\midrule
Token coverage & 90.9 & 94.6 \\
Uncovered images routed $\uparrow$ & 76.1 & 72.6 \\
Covered images misrouted $\downarrow$ & \textbf{3.4} & 7.7 \\
\midrule
Constrained only & 83.0 & 84.6 \\
With fallback & \textbf{86.0} & 82.1 \\
With oracle routing & \textbf{88.3} & 87.1 \\
\bottomrule
\end{tabular}
\caption{Routing analysis on the original model (\%, all 5{,}389 images). The bottom block reports token accuracy. Oracle routing uses the constrained output for covered words and the greedy output otherwise.}
\label{tab:routing}
\end{table}

To separate the quality of the routing rule from the choice of lexicon, we also evaluate oracle routing, which uses the constrained output whenever the gold word is in the lexicon and the greedy output otherwise. Even under oracle routing, the dictionary reaches a higher token accuracy than \Nstar{} (88.3\% vs.\ 87.1\%). In other words, once uncovered words can be passed to the open decoder, the additional coverage of a generated lexicon no longer compensates for its lower precision. The remaining 2.3-point gap between actual and oracle routing for the dictionary indicates how much a better routing rule could still gain.

\section{Conclusion}

We studied how the lexicon affects constrained decoding for low-resource Manchu OCR, where almost all test words are unseen in training. By decomposing accuracy into lexicon coverage and in-lexicon precision, we showed that coverage is a property of the lexicon alone, while precision depends on both the recognizer and the size of the lexicon. Expanding the dictionary with ranked root--suffix combinations raises coverage substantially, with suffix-chain frequency as the key ranking signal, but it lowers precision where the dictionary already covers most words. Confidence fallback provides the missing coverage at a much smaller cost, and a dictionary with fallback outperforms a generated lexicon ten times its size, even under oracle routing. We therefore recommend a precise lexicon combined with a validated fallback as the default for agglutinative low-resource scripts, with lexicon expansion preferred only when the recognizer is strong enough to keep its precision cost small. In future work, we plan to test whether these findings extend to other agglutinative scripts such as traditional Mongolian.

\section*{Limitations}

Our study covers a single language and script, and we have not verified that the findings hold for other agglutinative scripts. The Transformer and the training-fraction models were trained with one seed each, so their results do not reflect seed variance. The character language model is a trigram trained only on training transcriptions, and a stronger language model could narrow the gap between open and constrained decoding. The fallback threshold was selected from a coarse grid of six values, and the oracle analysis covers token accuracy of the original model only. The three newly trained seeds use a different batch configuration from the original model. Our explanation of the precision cost rests on weak correlational evidence. Finally, the test tiers depend on a dictionary-based segmenter that cannot analyse 158 of the test types. Because the split holds out word forms rather than documents, test images share publication sources, and possibly pages, with training images; our results therefore measure generalization to unseen word forms, not to unseen collections or writers.

\section*{Ethics Statement}

This work is intended to support the retrieval, transcription and collation of historical Manchu documents. The images are taken from previously published historical materials (Appendix~\ref{app:data}), which may contain perspectives specific to their period. OCR outputs should not be treated as authoritative transcriptions: when they are used for archival organization or collation, researchers with expertise in Manchu should review them against the originals, especially historical spellings, word boundaries and out-of-lexicon outputs. Word segmentation, transliteration and quality review were carried out by Manchu experts, all adult researchers who were informed of the purpose and intended use of the data and participated voluntarily; the protocol received institutional ethics approval. Copyright-restricted source scans and image derivatives will not be redistributed. We plan to release code under the Apache 2.0 license and research-created transliterations, splits and metadata under CC BY 4.0, subject to the rights of the source materials.

\nocite{chaoke2025}
\bibliography{references}

\appendix

\section{Data Sources}
\label{app:data}

Table~\ref{tab:sources} lists the source publications. Manchu50000 contains printed text with comparatively regular glyph forms, whereas the three handwritten collections show more writer variation, stroke deformation and ink variation.

\begin{table*}[h]
\centering\small
\begin{tabular}{llp{5.0cm}p{3.4cm}r}
\toprule
Dataset & Script & Source publication & Publication information & Test images \\
\midrule
Manchu50000 & Printed & \textit{Taizong}, vols.\ 1--4; translated and annotated by the Manbun R\=ot\=o Kenky\=ukai & Tokyo: T\=oy\=o Bunko, 1959--1963 & 3{,}940 \\
MengGuTangDang & Handwritten & \textit{Qing Neige Menggu Tangdang}; compiled by the First Historical Archives of China and the School of Mongolian Studies, Inner Mongolia University & Hohhot: Inner Mongolia People's Publishing House, 2005 & 1{,}037 \\
SuiJunJiXing & Handwritten & \textit{Suijun Jixing Yizhu}; written by Zeng Shou, translated and annotated by Ji Yonghai & Beijing: Central University for Nationalities Press, 1987 & 253 \\
Junjichu & Handwritten & \textit{Junjichu Yongzheng Chao Manwen Yifudang Yibian}; compiled by the First Historical Archives of China & Beijing: The Commercial Press, 2021 & 159 \\
\bottomrule
\end{tabular}
\caption{Source publications of the word images.}
\label{tab:sources}
\end{table*}

\section{Compute}
\label{app:compute}

Each CRNN run trains for 30 epochs on a single RTX 4090 GPU in about 12--13 minutes. The Transformer run takes about 8.5 minutes, and the runs on 25\% and 50\% of the training data take about 4 and 7 minutes, respectively.

\section{Additional Figures}

\begin{figure*}[h]
\centering
\includegraphics[width=0.75\textwidth]{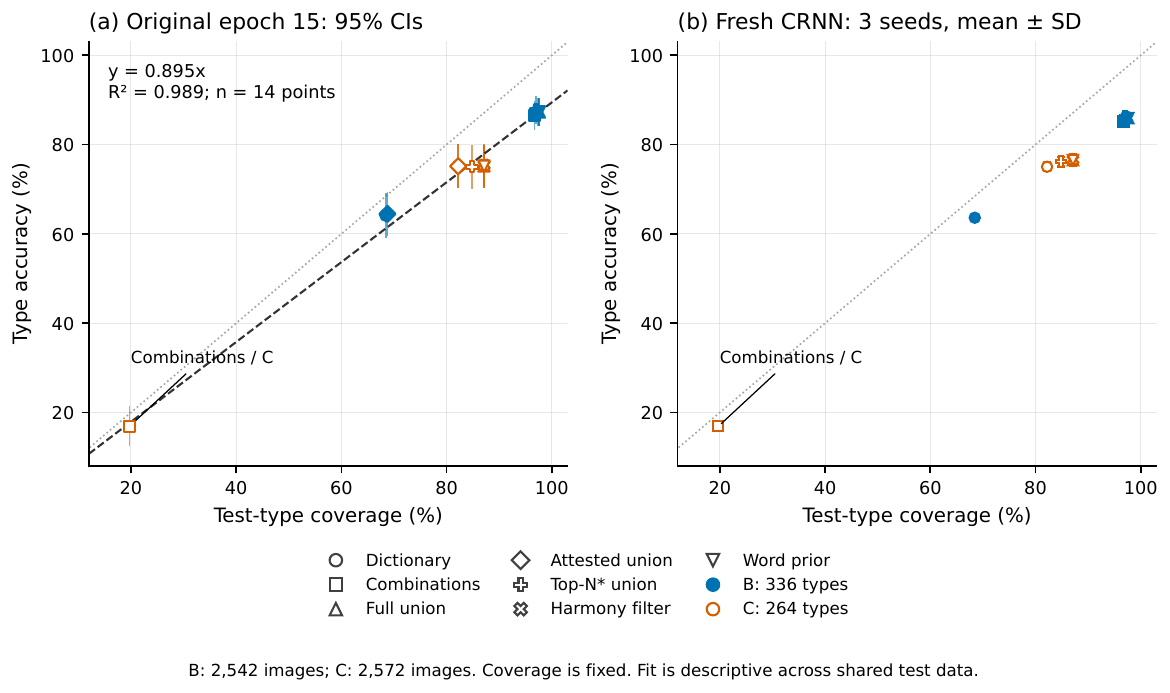}
\caption{Type accuracy against type coverage for tiers B and C. (a) Original model with 95\% word-type cluster CIs; the dashed line is a descriptive fit through the origin ($y=0.895x$, $R^2=0.989$, 14 points). (b) Three newly trained CRNN seeds, mean $\pm$ SD.}
\label{fig:fit}
\end{figure*}

\begin{figure*}[h]
\centering
\includegraphics[width=0.75\textwidth]{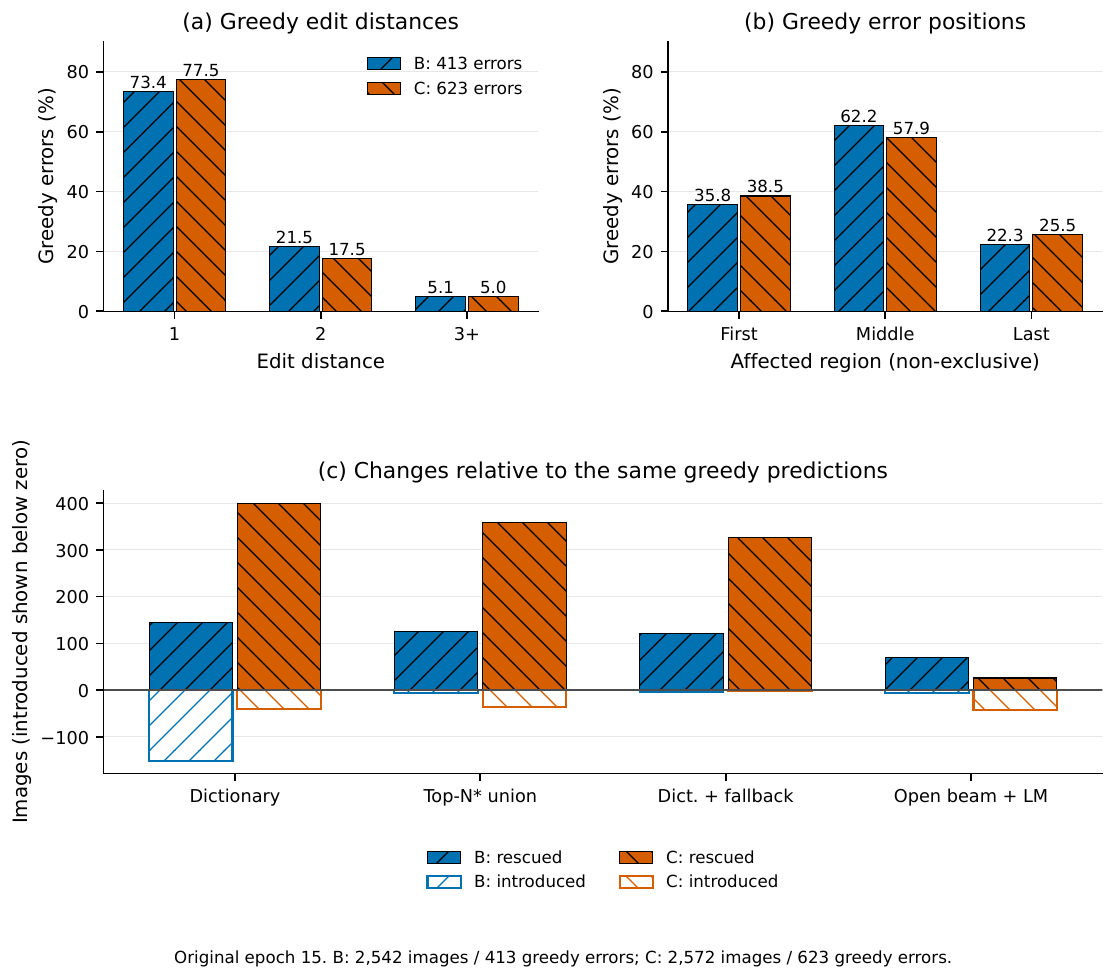}
\caption{Greedy CTC errors on tiers B and C: (a) edit distance to the gold word; (b) affected character region.}
\label{fig:errors}
\end{figure*}

\begin{figure*}[h]
\centering
\includegraphics[width=0.8\textwidth]{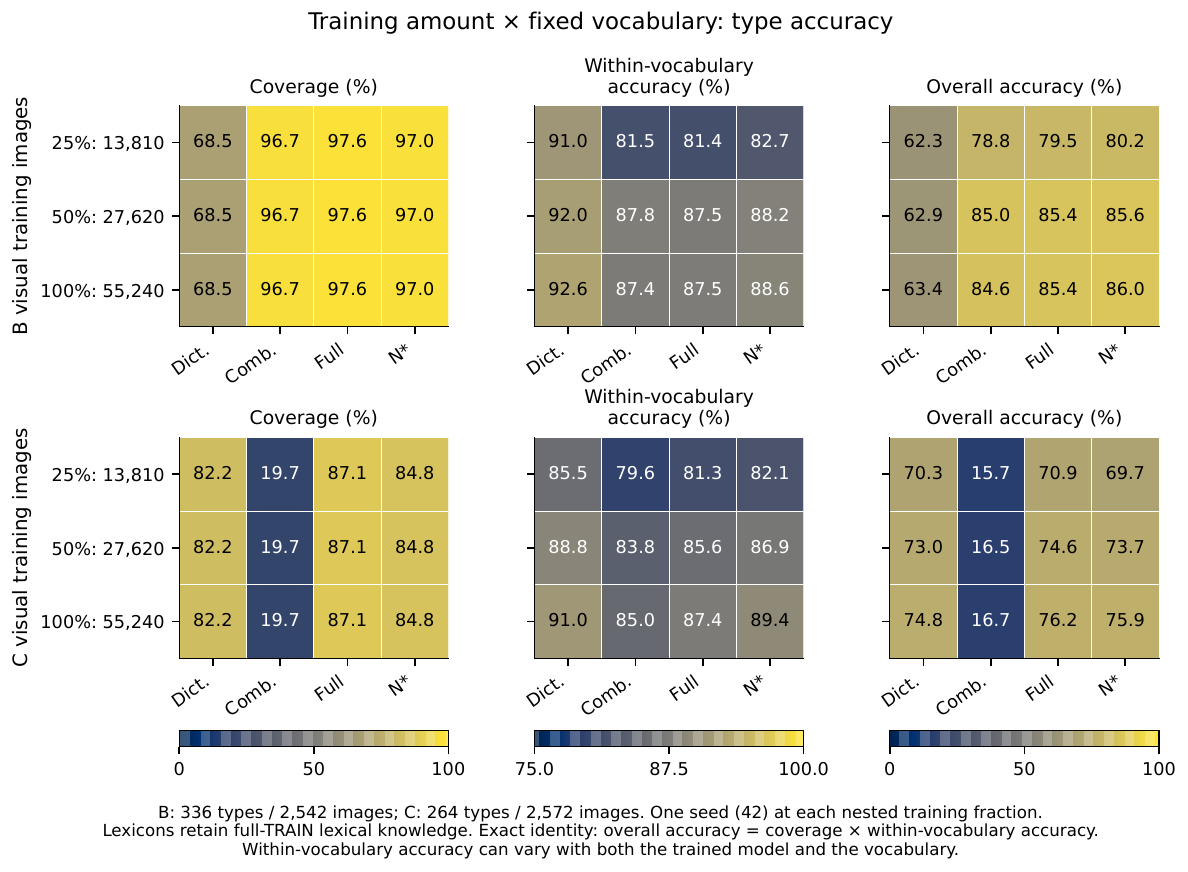}
\caption{Training amount $\times$ lexicon: coverage, in-lexicon precision and type accuracy for tiers B and C.}
\label{fig:crossfig}
\end{figure*}

\begin{figure*}[h]
\centering
\includegraphics[width=0.8\textwidth]{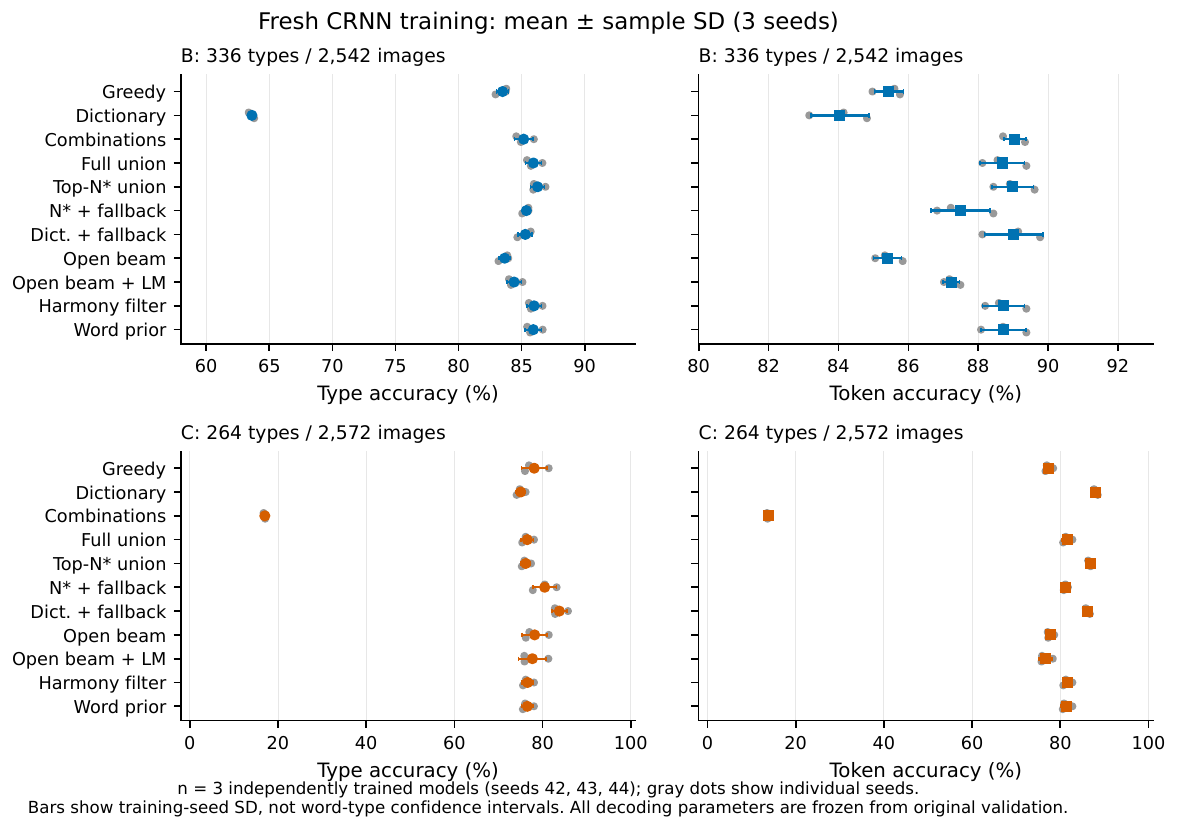}
\caption{All systems on tiers B and C for three CRNN training seeds (mean $\pm$ SD; grey dots are individual seeds).}
\label{fig:seeds}
\end{figure*}

\begin{figure*}[h]
\centering
\includegraphics[width=0.8\textwidth]{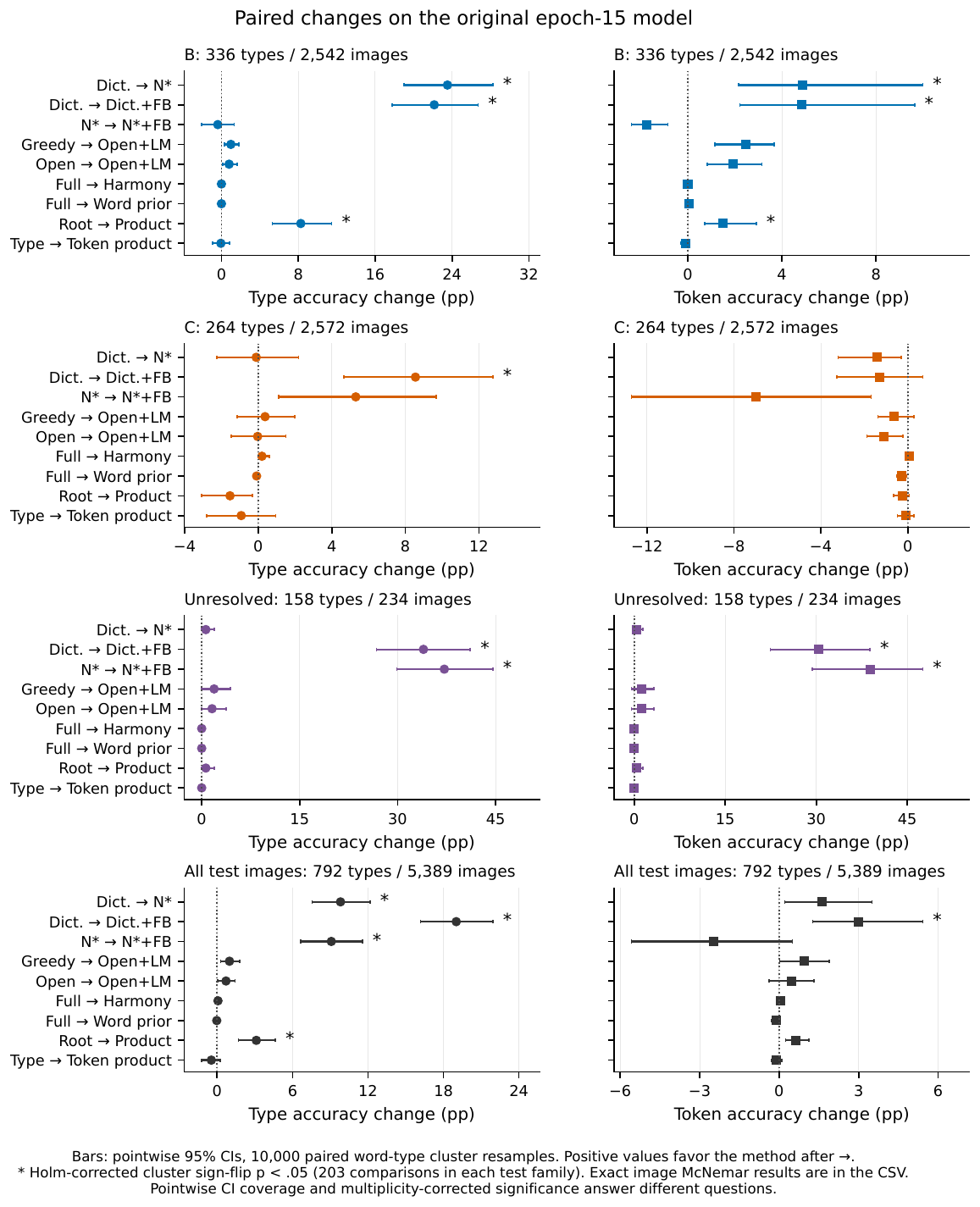}
\caption{Paired differences between systems with 95\% word-type cluster bootstrap CIs.}
\label{fig:paired}
\end{figure*}

\begin{figure*}[h]
\centering
\includegraphics[width=0.8\textwidth]{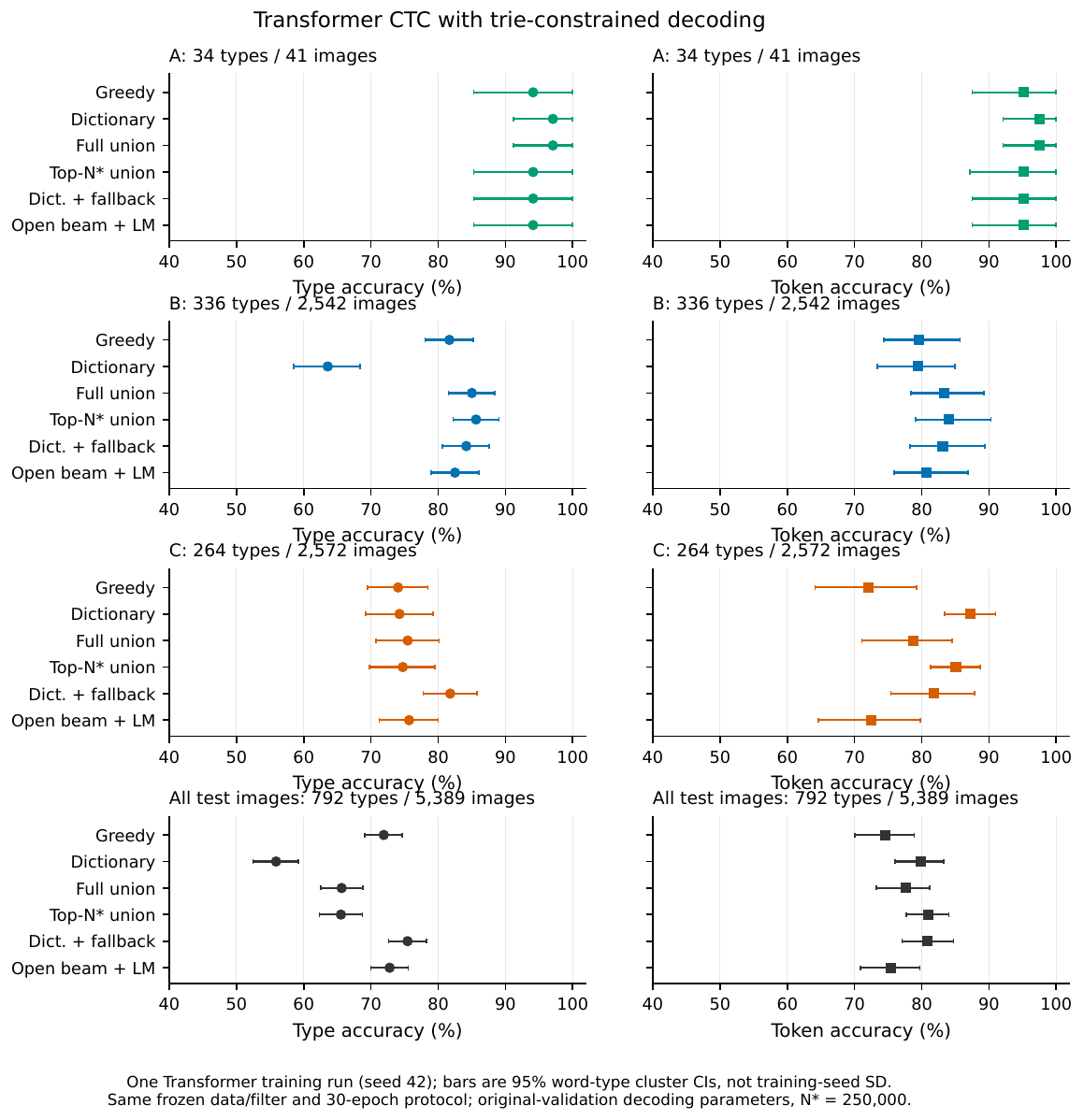}
\caption{Transformer recognizer, one seed, with 95\% word-type cluster bootstrap CIs.}
\label{fig:transformer}
\end{figure*}

\begin{figure*}[h]
\centering
\includegraphics[width=0.7\textwidth]{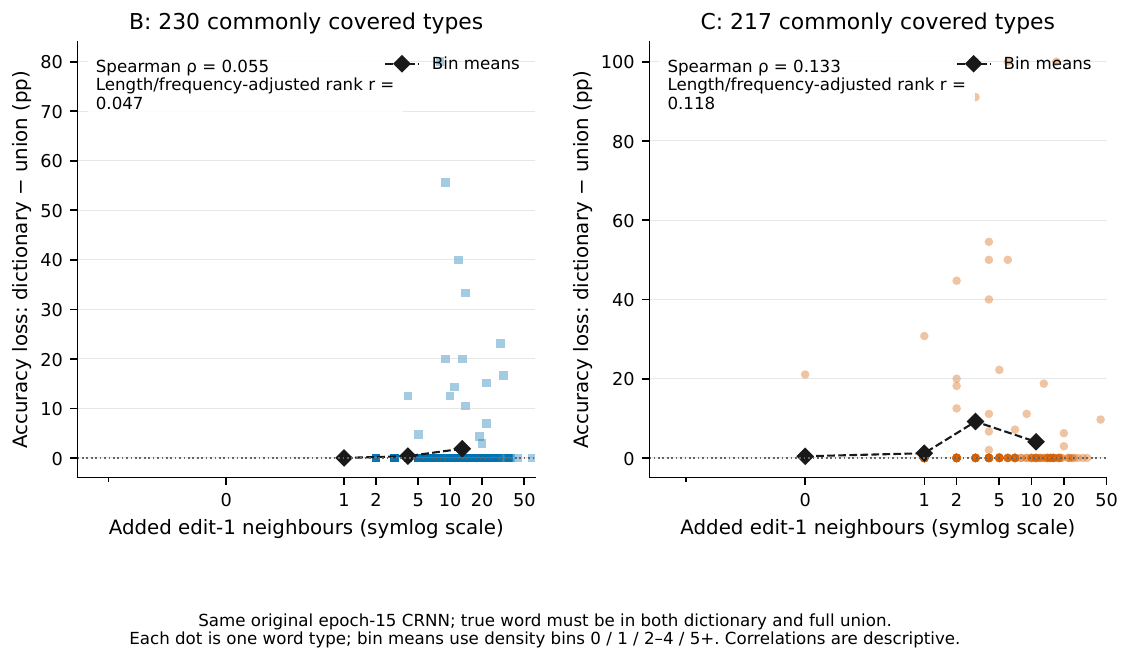}
\caption{Added edit-distance-1 neighbour density against per-type accuracy loss when moving from the dictionary to the full union.}
\label{fig:density}
\end{figure*}

\begin{figure*}[h]
\centering
\includegraphics[width=0.8\textwidth]{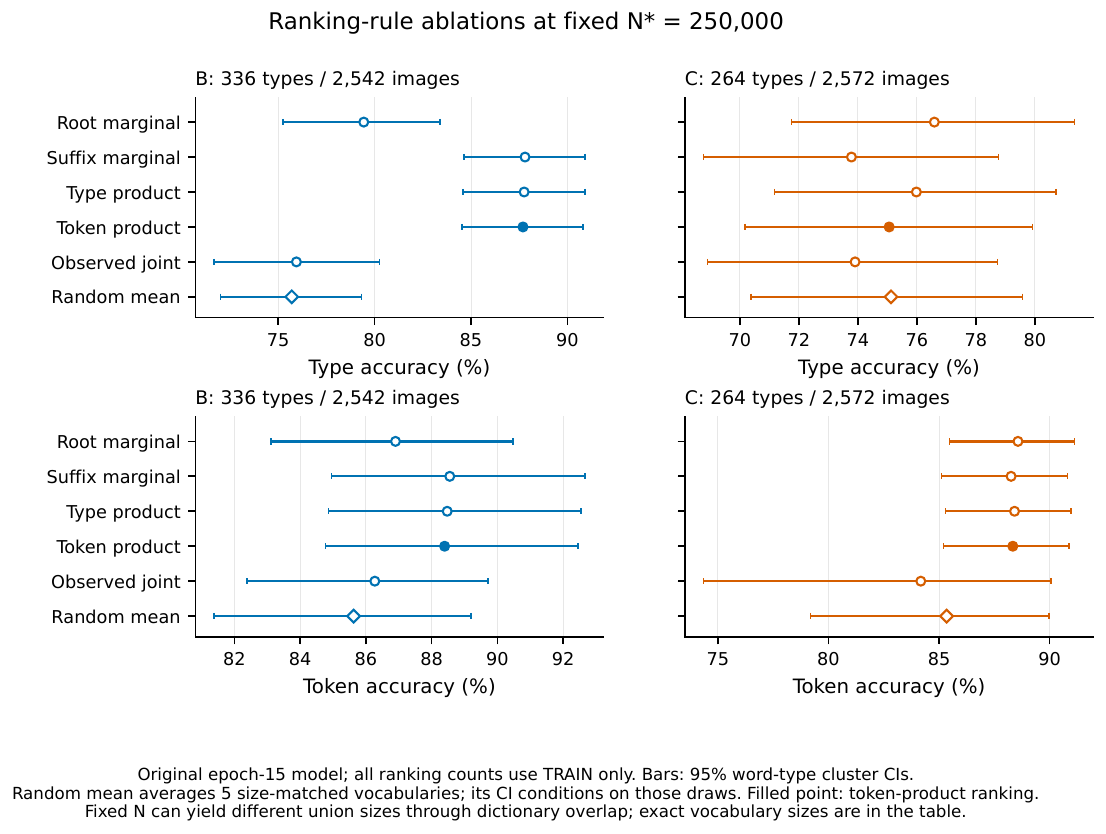}
\caption{Ranking rules for generated candidates with 95\% CIs.}
\label{fig:rankfig}
\end{figure*}

\begin{figure*}[h]
\centering
\includegraphics[width=0.7\textwidth]{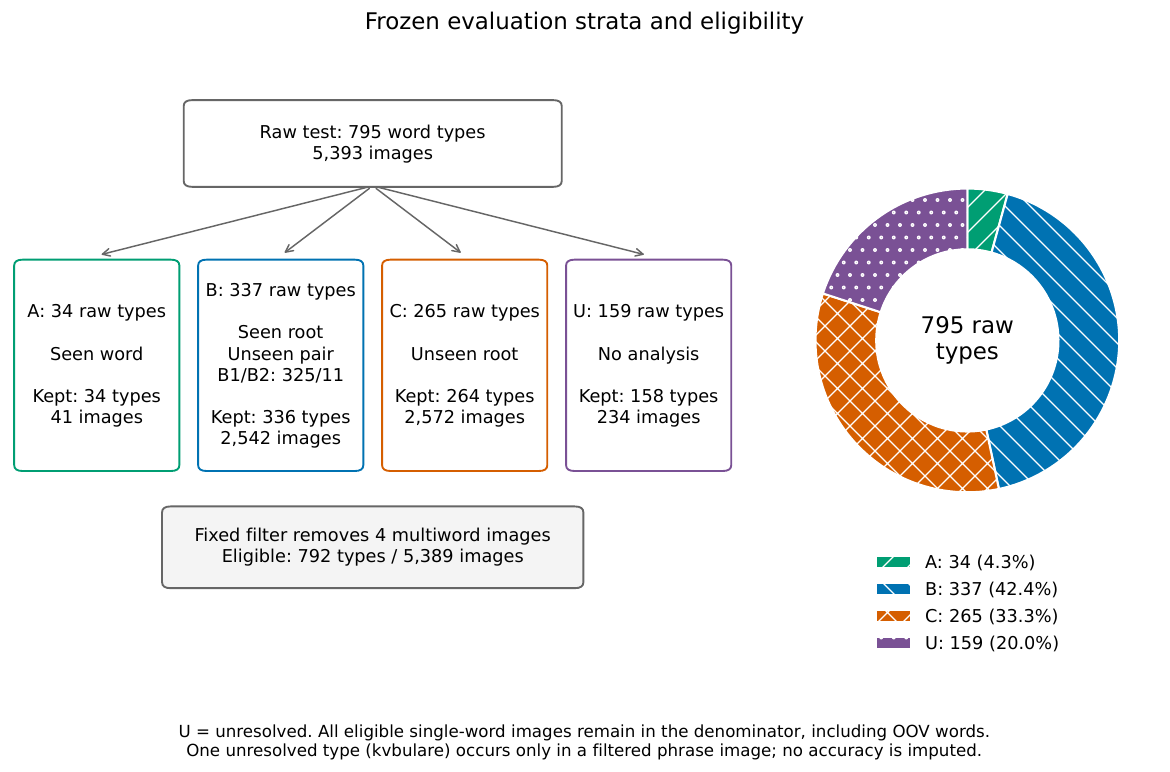}
\caption{Construction of the test tiers.}
\label{fig:protocol}
\end{figure*}

\end{document}